\documentclass[10pt,twocolumn,letterpaper]{article}

\usepackage{cvpr}              

\usepackage[dvipsnames]{xcolor}

\usepackage{amsmath}
\usepackage{graphicx}
\usepackage{amsmath,amssymb,latexsym,epsfig,amsthm}
\usepackage{float} 
\usepackage{subcaption}
\usepackage{algorithm}
\usepackage{algpseudocode}
\usepackage{multirow}

\definecolor{cvprblue}{rgb}{0.21,0.49,0.74}
\usepackage[pagebackref,breaklinks,colorlinks,citecolor=cvprblue]{hyperref}

\def\paperID{*****} 
\def\confName{CVPR}
\def\confYear{2024}

\title{Applying ViT in Generalized Few-shot Semantic Segmentation}

\author{Liyuan Geng \textsuperscript{*}\\
NYU Shanghai\\
{\tt\small lg3490@nyu.edu}
\and
Jinhong Xia \textsuperscript{*}\\
NYU Shanghai\\
{\tt\small jx2314@nyu.edu}
\and
Yuanhe Guo\\
NYU Shanghai\\
{\tt\small yg2709@nyu.edu}
}

\begin{document}
\maketitle

\begin{abstract}
    This paper explores the capability of ViT-based models under the generalized few-shot semantic segmentation (GFSS) framework. We conduct experiments with various combinations of backbone models, including ResNets and pretrained Vision Transformer (ViT)-based models, along with decoders featuring a linear classifier, UPerNet, and Mask Transformer. The structure made of DINOv2 and linear classifier takes the lead on popular few-shot segmentation bench mark PASCAL-$5^i$, substantially outperforming the best of ResNet structure by $116\%$ in one-shot scenario. We demonstrate the great potential of large pretrained ViT-based model on GFSS task, and expect further improvement on testing benchmarks. However, a potential caveat is that when applying pure ViT-based model and large scale ViT decoder, the model is easy to overfit. Our code is available at \href{https://github.com/LGNYU/ViTSeg}{\texttt{https://github.com/LGNYU/ViTSeg}}.
\end{abstract}
\begingroup
\renewcommand\thefootnote{}
\footnotetext{* Equal contribution.}
\endgroup
\section{Introduction}
\label{sec:introduction}

Segmenting image based on semantic understanding has achieved remarkable performance in recent years, thanks to deep learning methods. With an image input, these models~\cite{7913730, Zhao_2017_CVPR} are capable of predicting which classes each pixel belongs to. However, these models are usually limited by the scale of training data. Most of standard segmentation methods could only predict a fixed set of classes predefined by the training data. To reach a satisfying performance, each class in the training data often requires hundreds of well labeled images. When utilizing the model in novel classes, obtaining annotations is labor-intensive, and thus limits the model scalability. 

Few-shot segmentation (FSS) are proposed as a method to quickly adopt standard segmentation models to previous unseen classes with only limited annotated data~\cite{9577299}. With the goal of predicting novel classes, these paradigms train models using sufficient labeled data on \textit{base} classes during training stage, and then a few instances of \textit{novel} classes during inference stage. However, these methods fell short at tackling real-world scenarios due to limitations spotted in~\cite{tian2022generalized}. Primarily, FSS strongly assumes that classes in query images are fully covered in support images, resulting in costly manual efforts during data preparation. Further more, FSS models are only evaluated on novel classes separately, while the decline of the performance on base classes and the other novel classes is underestimated. 

Recently, a new task called Generalized Few-shot Segmentation (GFSS) emerges to handle these limitations. To be specific, this new pipeline doesn't require support images to have exactly the same classes as query images. Besides, models are evaluated by both base and novel classes, thus getting closer to the real-world use cases. The latest SOTA work~\cite{Hajimiri_2023_CVPR} manages to disentangle the training and testing phases, acquiring the ability to address arbitrary tests during testing. Although substantial achievements has been made, we identify points that worth further exploration and are not covered in existing literature. A more detailed discussion can be found in Sec.~\ref{sec:method}

\paragraph{Backbone architecture.} The idea of training a model with abundant data and fine-tune it on other tasks is exactly what large pretrained Visual Transformer (ViT)~\cite{dosovitskiy2021image} models expert in. Note that even though DIaM~\cite{Hajimiri_2023_CVPR} acquired impressive performance, their method is built upon ResNet~\cite{he2015deep} architecture. Thus, we wonder if ViT-based pretrained model could tackle GFSS task better than ResNet-based model. 

\paragraph{Loss Function.}  DIaM~\cite{Hajimiri_2023_CVPR}
introduced a sophisticated loss function to enhance performance on novel classes while preserving base knowledge. Nevertheless, tuning the augmented set of hyperparameters poses a challenge. Hence, we opt for a straightforward cross-entropy loss to maintain a fair comparison between different backbone architectures.

\paragraph{Contributions.} We experiment on multiple combinations of ViT-based pretrained models with different decoders, and compare them with ResNet-based models, under the paradigm of GFSS, and demonstrate the stunning capability of ViT-based models. Specifically:
\begin{itemize}
    \item We show that under the same dataset and training condition, ViT-based model DINO~\cite{caron2021emerging} and DINO v2~\cite{oquab2023dinov2} outperforms ResNet-based model. Meanwhile, DINO v2 has significant advantage over DINO in our few-shot segmentation task. 
    \item In view of our observations, Mask Transformer~\cite{NEURIPS2021_950a4152} is more likely to overfit than linear classifier when connected to DINOv2.
\end{itemize}

\section{Related Work}
\label{sec:related_work}

\paragraph{Few-shot segmentation.} Few-shot semantic segmentation (FSS) extends the successful few-shot learning paradigm~\cite{Cai_2018_CVPR, ravi2017optimization} to the task of semantic segmentation~\cite{Long_2015_CVPR, 7913730, Zhao_2017_CVPR}. Given only a few support examples, models are expected to predict dense pixel labels on new classes. Foundational FSS approaches leveraged a dual-branch framework, where one branch generated prototypes from support images, and the other branch took query images and these prototypes into a segmentor to predict masks~\cite{dong2018few, shaban2017oneshot}. Subsequent literature focused on more efficiently mining information from support images to guide the segmentation of query images. For instance, some divide one support image into multiple dynamically allocated prototypes to yield more information~\cite{Li_2021_CVPR, 10.1007/978-3-030-58598-3_45}. Besides, graph attention mechanism was utilized to establish correspondence between support and query images~\cite{10.1007/978-3-030-58601-0_43, Zhang_2019_ICCV}. More recently, a bunch of works leveraged visual transformers to fulfill the same goal.~\cite{NEURIPS2022_f7fef21d, Lu_2021_ICCV}. 

\paragraph{Generalized Few-shot segmentation.} A new setting, termed as generalized few-shot semantic segmentation (GFSS), emerges to resolve some of the shortcomings of FSS. Compared with FSS models, GFSS takes a set of support images for all novel classes, and predicts both potential base and novel classes in query images. CAPL~\cite{tian2022generalized} marked the first trial on handling GFSS. Their framework consists of two context-aware components to dynamically adapt both base and novel prototypes. To further improve model performance on GFSS task, DIaM~\cite{Hajimiri_2023_CVPR} assembled multiple optimization goals into one loss function that maximizes mutual information between learned features and their predictions, while preserving base knowledge.

\paragraph{Visual Transformer.} In recent days, we have witnessed the success of transformer architecture~\cite{NIPS2017_3f5ee243} in natural language processing. This architecture made up of self-attention and feed-forward layers also acquires stunning results in computer vision. Visual transformer (ViT)~\cite{dosovitskiy2021image} first introduced transformer to vision tasks through tokenizing patches split from an image. Transformer-based approaches covers a variety of tasks, including semantic segmentation~\cite{Zheng_2021_CVPR}, object detection~\cite{10.1007/978-3-030-58452-8_13}, multiple object tracking~\cite{sun2021transtrack}. Given the transformer architecture's remarkable performance on large scale dataset, a group of works explored in pretraining ViT-based models for learning task-agnostic representations. DINO proposed a simple yet effective self-supervised learning framework~\cite{caron2021emerging}. This was further enhanced by DINOv2~\cite{oquab2023dinov2}, which was pretrained on large curated data with no supervision. 

\paragraph{Visual Transformer in Segmentation. } Recently, with the rising of Vision Transformer, some applications of Vision Transformer to the semantic segmentation task has emerged. SegFormer~\cite{xie2021segformer} introduces an architecture that consists of a hierarchical Transformer encoder to generate high-resolution
coarse features and low-resolution fine features and a lightweight All-MLP decoder to fuse these
multi-level features to produce the final semantic segmentation mask, which shows high efficiency and performance. Besides, Segmentor~\cite{strudel2021segmenter} is based on a fully transformer-based encoder-decoder architecture mapping a sequence of patch embeddings to pixel-level class annotations. In particular, the architecture uses Mask Transformer~\cite{cheng2022maskedattention} as its decoder, which takes image patches and class embeddings as input. It is suitable for doing few shot learning, since we can initialize new class embeddings flexibly.
\section{Method}
\label{sec:method}

\begin{figure}[ht!]
    \centering
    \includegraphics[width=1\linewidth]{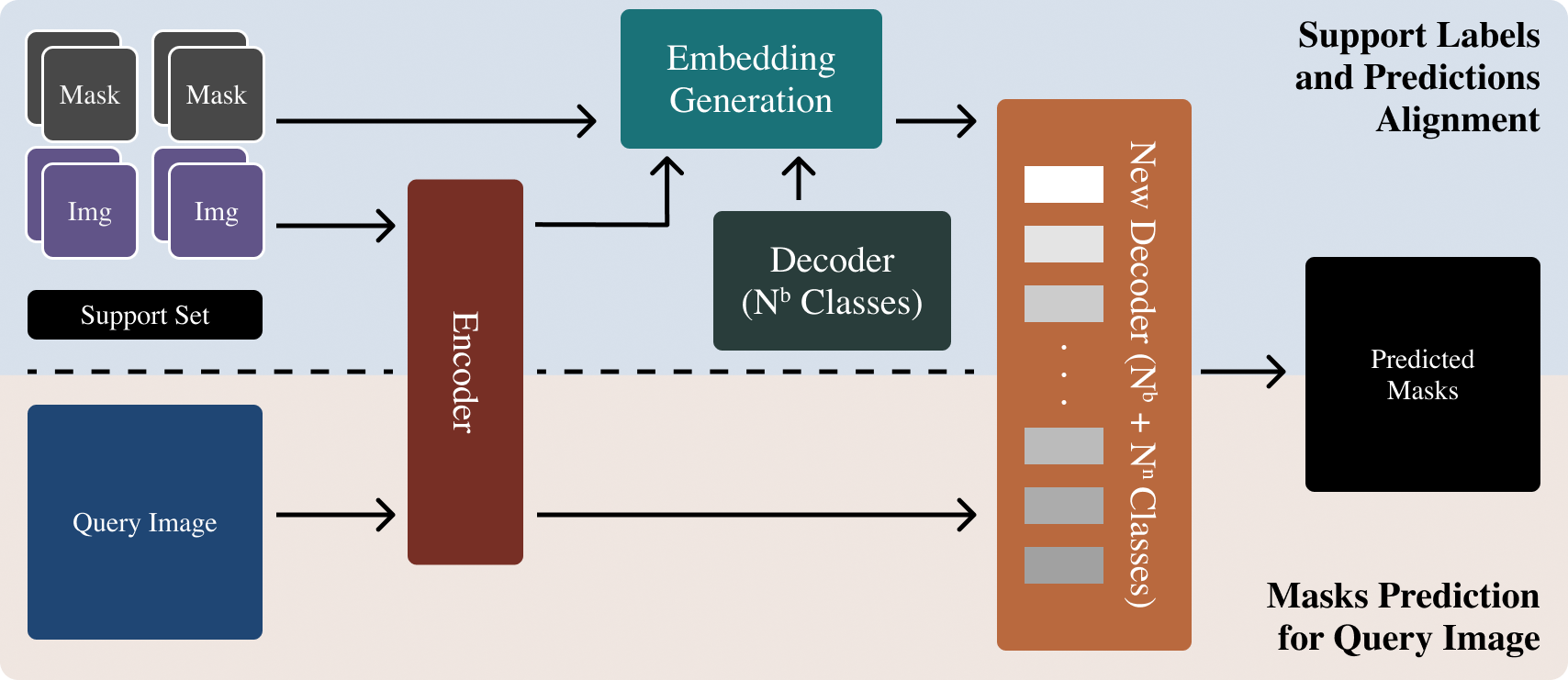}
    \caption{Overview of  our structure for generalized few-shot segmentation. The upper-half shows the process of aligning support labels and predictions, while the lower-half representing the mask prediction process for the query image.  The disentanglement between \textit{Encoder} and \textit{Decoder} enables us to experiment on different model combinations: \textit{ResNet34+Linear Classifier, ResNet34+UperNet, ResNet50+Linear Classifier, ResNet50+UperNet, DINO+Linear Classifier, DINO+Mask Transformer, DINOv2+Linear Classifier, DINOv2+Mask Transformer}. }
    \label{fig:framework}
\end{figure}

In light of the Generalized Few-shot Semantic Segmentation framework \cite{Hajimiri_2023_CVPR}, we divide the whole procedure of generalized FSS method into two stages, training and inference, and we will focus on inference stage since it is the core part of GFSS. We want to test different model architectires' ability to 1) generalize information from base classes 2) quickly converge in novel classes 3) maintain knowledge of base classes. 
\subsection{Base Training}
We devide the segmentation model into a feature extractor (encoder) $f_\phi$ and decoder, which outputs a segmentation map. We apply the standard training procedure during this stage, where we apply training and standard validation after each epoch, and choose the model with the best validation result , i.e. highest mIoU, as the model we will use as the initial model for the second stage. During this phase, the model is trained only to segment base class $C^b$, so the model can only predict $1 + |C^b|$ classes, i.e. the background and base classes. This is the general base training paradigm.

However, there are still minor differences in terms of different model architectures that we apply during this stage.

\subsubsection{ResNet}
For ResNet backbone, the training prcodure starts from ResNet 34 and ResNet 50 pretrained backbone, and the randomly initialized linear classifier and UPerNet~\cite{xiao2018unified}, which is very similar to the approach in the base training procedure in DIaM~\cite{Hajimiri_2023_CVPR}. In the base training stage, all parameters in the whole encoder-decoder architecture are trained. 
To help model better understand the image domain, we choose UPerNet as our decoder for ResNet. Inspired by human perception of visual information, it merges information of textures, parts, objects, and scenes together by directing different layers output to corresponding segmentation heads. This module takes the advantage of CNN in grabbing information of different scales and help the model understand the global and local information. For our training paradigm, we only keep the segmentation head for objects, and we freeze it in inference stage to avoid overfitting. 

\subsubsection{Vision Transformer}
For ViT backbone, we directly load DINO \cite{caron2021emerging} and DINOv2 \cite{oquab2023dinov2} as the pretrained backbone given the huge amount of parameters of Vision Transformer. During the base training, we freeze the pretrained backbone, and train the linear classifier and mask transformer as the decoder of the whole network, respectively.

\subsection{Inference}
 In the inference stage, similar to the approach in DIaM~\cite{Hajimiri_2023_CVPR}, given $|C^n|$ novel classes to recognize, we freeze the feature extractor $f_\phi$ and augment the decoder through various ways according to different model architectures. We augment the decoder by training it through support images, and we evaluate the few shot performance through query image. However, in the inference stage, we do not require object of base classes to be labeled, so we adjust the label a little bit to align the support labels with predictions from the model.
\paragraph{Aligning support labels and predictions.} In order to address the misalignment between the support image labels and the predictions from the model,  we adjust the predictions from the model using the following pseudocode.

\begin{algorithm}[H]
\caption{adjustPrediction(probs)}
\label{adjustPrediction}
\begin{algorithmic}

\State newBg = probs[ : , 0:$|C_b |$].sum(dim=1, keepdim=True) 

\State newBase = torch.zeros\_like(probs[ : , 1 : $|C_b |$]) 

\State newNovel = probs[ : , $|C_b |$] 

\State newProbs = torch.cat((newBg,newBase,newNovel),dim=1)
\State return newProbs 
\end{algorithmic}
\end{algorithm}

As it is shown in Algorithm \ref{adjustPrediction}, we set the background of the adjusted prediction as the sum of the dimension of the original prediction from 0 to $|C_b |$. We set the base dimension of new prediction as 0s, and we keep the novel class dimension. Then we concatenate all dimensions to get the adjusted prediction. The adjusted model prediction for each pixel can be demonstrated as the following: 
\begin{align*}
    \bigg[\sum_{k=0}^{\left|\mathcal{C}^b\right|} p_k, \overbrace{0, \ldots, 0}^{\left|\mathcal{C}^b\right| \text { times }}, p_{\left|\mathcal{C}^b\right|+1}, \ldots, p_{\left|\mathcal{C}^b\right|+\left|\mathcal{C}^n\right|}\bigg]^{\top}
\end{align*}

After applying this adjustment to the prediction, we can use cross entropy loss to the prediction and the ground truth label. However, for different model architectures, in particular, for different decoders, there are some differences in terms of the implementation of augmenting the decoder.

\subsubsection{Linear Classifier / UPerNet}
Similar to DIaM~\cite{Hajimiri_2023_CVPR}, we augment the pre-trained classifier $\theta_b$ with novel prototype $\theta_n$, and we set the $[\theta_b; \theta_n]$ as our final classifier. Then we use the support image to train the model. The parameters in UPerNet~\cite{xiao2018unified} will be completely freezed during the inference stage.

\subsubsection{Mask Transformer}
In light of the architecture of Segmentor~\cite{strudel2021segmenter}, we use Mask Transformer~\cite{cheng2022maskedattention} as the decoder, and the way we augment the decoder is to randomly initialize $|C_n |$ class embeddings. Then those novel class embeddings combined with the original image patches and base class embeddings will be fed into the Mask Transformer to get the final segmentation map. During the inference stage, the class embeddings and the layernorm parameters in the Mask Transformer are treated as parameters when training with support images.

\section{Experiment}
\label{sec:experiment}

\subsection{Experiment setting}
\paragraph{Datasets.} To compare and evaluate different models, we use a well-known semantic segmentation dataset: PASCAL-$5^i$~\cite{shaban2017oneshot}. The dataset is build on PASCAL VOC 2012~\cite{pascal-voc-2012} (containing 20 semantic classes) with additional annotations from SDS~\cite{hariharan2014simultaneous}. We set the first 5 classes as the novel class, and the rest of 15 classes as the base class.

\paragraph{Evaluation Protocol.} To evaluate the model performance under GFSS settings of different model architectures, we track the mean intersection-over-union (mIoU) over the classes. Although some prior works track the mIoU over all classes as the overall score~\cite{lang2022learning}, in our GFSS settings, we will track the mIoU over base classes and novel classes, respectively, in order to test models' ability to quickly learn novel classes and its generalized knowledge of base classes to avoid high decreases of mIo U over base classes after adapting to novel classes. In our tables, \textit{Base} and \textit{Novel} refer to mIoU over bases classes and novel classes, respectively. In accordance with the implementation in \cite{lang2022learning}, the mIoU are averaged through 5 independent runs. 

\paragraph{Implementation details.} In the base training stage, we train the model using vanilla cross entropy loss for 100 epochs, and validate after each epoch. We save the model which has the highest validation mIoU. In the base training process, we mostly follow the DIaM settings~\cite{Hajimiri_2023_CVPR}, and hence the batch size is 8, and the SGD optimizer is used with an initial learning rate $2.5 \times10^{-4}$, momentum 0.9, and weight decay $10^{-4}$. In the inference stage, for each run, we go through all suitable query images, and, similar to the implementation in DIaM~\cite{Hajimiri_2023_CVPR}, for each query image, we load the base weight and pretrained weight of the parameters and randomly initialize the newly added parameters associated with the few shot learning. Then for each query image, we train the decoder with $|C_n| \times shots$ support images, and those $|C_n| \times shots$ of support images are different for each query image. For example, we evaluate the result using PASCAL-$5^i$ for 5 shots, then for each query image, we have $5 \times 5 $ support images to train the decoder. For this phase, SGD optimizer is used with learning rate $1.25 \times 10^{-3}$ and the cross entropy loss is optimized for 300 iterations.

\begin{table*}[ht!]
    \centering
\begin{tabular}{|c|c|c|c|c|c|c|c|c|c}
\hline \multirow{3}{*}{ Encoder } & \multirow{3}{*}{ Decoder } & \multicolumn{7}{|c|}{ PASCAL } \\
\hline & & \multicolumn{1}{|c|}{ Base Training }& \multicolumn{3}{|c|}{ 1-Shot } & \multicolumn{3}{|c|}{ 5-Shot } \\
\hline & & & Base & Novel & Mean & Base & Novel & Mean \\
\hline ResNet 34~\cite{he2015deep} & Linear &0.5539&  0.2487& 0.0846 & 0.1667 & 0.1546 & 0.0593 & 0.1070\\
\hline ResNet 34~\cite{he2015deep} & UperNet~\cite{xiao2018unified} & 0.6736&0.4650 & 0.0722 & 0.2686 & 0.4735& 0.0635& 0.2685 \\
\hline ResNet 50~\cite{he2015deep} & Linear &0.5939& 0.3763 & 0.1260 & 0.2512&0.1705
&0.0680 &0.1193  \\
\hline ResNet 50~\cite{he2015deep} & UperNet~\cite{xiao2018unified} &0.7038& 0.4739 &  0.0840& 0.2790 &0.4635&0.0726&0.2681 \\
\hline DINO~\cite{caron2021emerging} & Linear & 0.5747 & 
 0.1240 & 0.1332 & 0.1286 & 0.1883 & 0.1777 & 0.1830 \\
\hline DINO~\cite{caron2021emerging} & Mask Transformer~\cite{cheng2022maskedattention} & 0.6803 & 0.6378 & 0.0806 & 0.3592 & 0.6437 & 0.2512 & 0.4475 \\
\hline DINOv2~\cite{oquab2023dinov2} & Linear & 0.7606 
& 0.7246 & \textbf{0.4799} & \textbf{0.6022} & 0.6857 & \textbf{0.5220} & 0.6039 \\
\hline DINOv2~\cite{oquab2023dinov2} & Mask Transformer~\cite{cheng2022maskedattention} & \textbf{0.7987} & \textbf{0.8171} & 0.2949 & 0.5560 & \textbf{0.8217} & 0.4598 & \textbf{0.6407} \\
\hline 

\end{tabular}

\caption{Quantitative results on PASCAL- $5^i$ for different encoder-decoder architectures.}
\label{tab:quantitative_eva}
\vspace{-10pt}
\end{table*}

\subsection{Experiment results and analysis}

\begin{figure*}[ht!]
    \centering
    \includegraphics[width=0.9\textwidth]{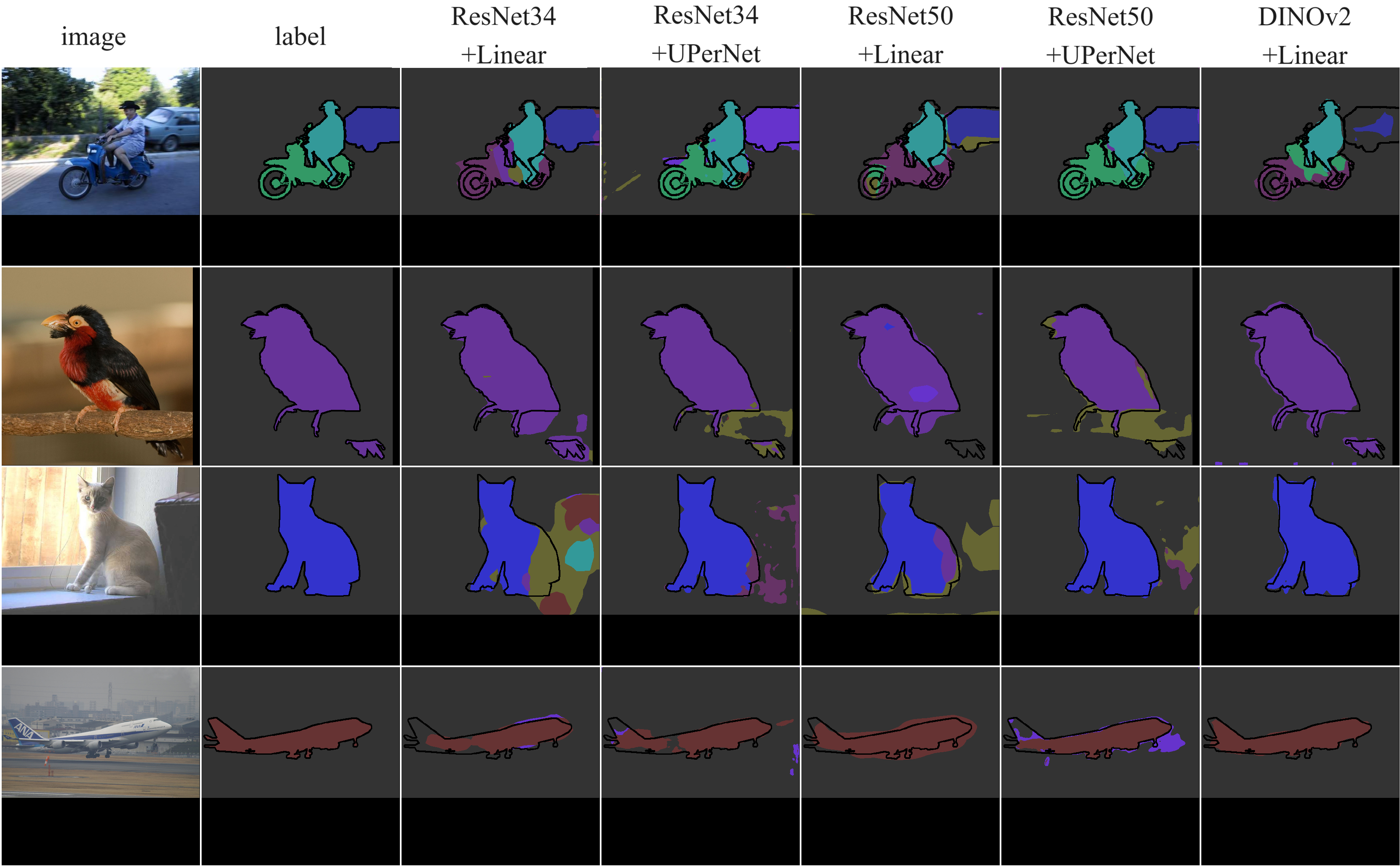}
    \caption{Qualitative results on 1-shot inference on PASCAL$-5^0$ for different encoder-decoder architectures. Models learn from support sets on novel classes \textit{airplane, bicycle, bird, boat, bottle}. Models will predict base and novel classes later.}
    \label{fig:qualitative}
\end{figure*}

Based on the result in Table \ref{tab:quantitative_eva}, we can make some comparisons across different encoder-decoder model architectures.

\paragraph{DINOv2 is a good few shot learner.} As is shown in the table, when we use DINOv2~\cite{oquab2023dinov2} as the backbone, whenever it is 1-shot or 5-shot setting, the whole model significantly outforms ResNet and DINO. Therefore, it is clear that DINOv2 is a good few shot learner. 

\paragraph{UPerNet preserves information on base classes and hinders few-shot learning.}
Since we don't apply any training strategy to inference stage, ResNet architectures ~\cite{he2015deep} do not have an ideal performance. However, by running the same learning protocols on different scale of ResNet, we can analyze the relation between model structures and its performance under GFSS paradigm. 

In the base training stage, ResNet-50 outperforms ResNet-34~\cite{he2015deep}, and the use of UPerNet~\cite{xiao2018unified} as the decoder significantly enhances performance compared to a linear classifier. This suggests that with sufficient training data, models with larger capacity and more complex architectures achieve superior results. During the inference stage, ResNet-50 continues to outperform ResNet-34 in both base and novel classes~\cite{he2015deep}. Although models with linear modules show a marked decrease in performance on base classes, they adapt better to novel classes than those using UPerNet. This observation indicates that the UPerNet~\cite{xiao2018unified} module, while effectively preserving information about base classes during the base training stage, struggles to generalize to novel classes during inference.

\begin{figure}[ht!]
    \centering
    \includegraphics[width=1\columnwidth]
    {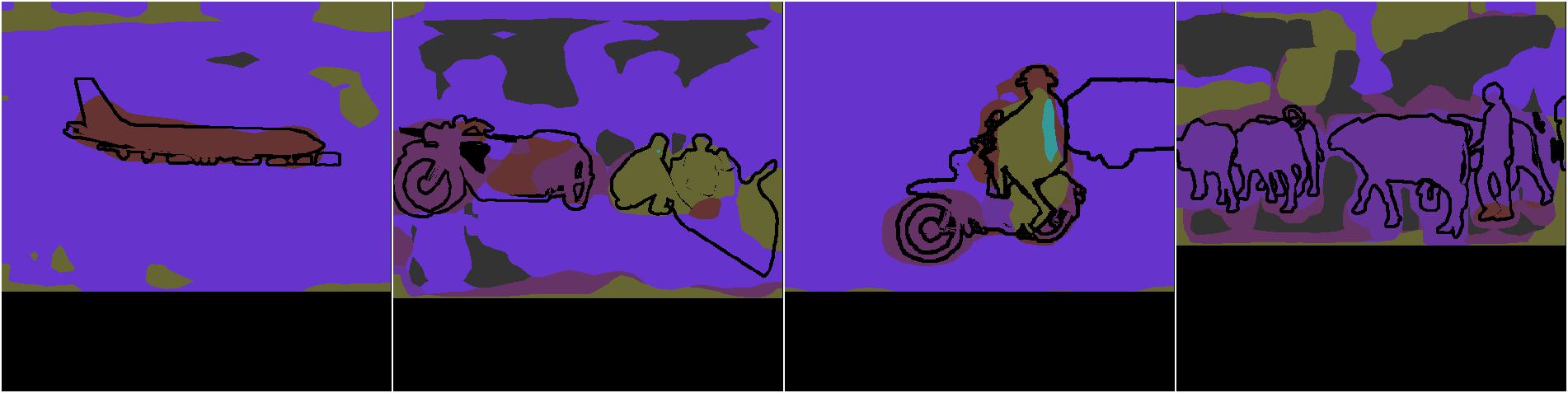}
    \caption{Example of results on 5-shot inference on PASCAL$-5^0$ using ResNet34 with linear classifier. Models mistake background pixels as novel classes.}
    \label{fig:5shot}
\vspace{-10pt}
\end{figure}

A counter-intuitive result is that ResNet backbones~\cite{he2015deep} perform worse in 5-shot learning compared to 1-shot learning. This issue arises because we do not impose any restrictions on the model when predicting background pixels as other classes, leading to the incorporation of incorrect information from novel classes and, consequently, a decline in performance. As illustrated in Figure~\ref{fig:5shot}, the model frequently misclassifies background pixels as novel classes. Based on these visualization results, we conclude that ResNet~\cite{he2015deep} struggles to generalize semantic information effectively across images and tends to learn incorrect information when exposed to more, yet insufficient, training samples.

\paragraph{Applying Mask Transformer as the decoder is easy to be overfitting in the GFSS setting.} In the table, a noticable trend is that no matter we use DINO~\cite{caron2021emerging} or DINOv2~\cite{oquab2023dinov2} as the backbone, the model performance of using linear classifier is much better than using mask transformer as the decoder over novel classes, while using mask transformer is better in terms of preserving knowledge of base classes. One possible explanation is that mask transformer overfits because  only few support images are used to train the novel class for each query image. The explanation can be supported by the fact that, in the DINOv2~\cite{oquab2023dinov2} + Mask Transformer~\cite{strudel2021segmenter} settings, 5-shot mIoU is significantly higher than 1-shot mIoU, since more support images mitigate the overfitting effect. 
\section{Conclusion}
\label{sec:conclusion}
We explore the capability of ViT-based models under the generalized few-shot semantic segmentation framework. We apply several combinations of encoder and decoder architectures, and our GFSS setting is based on DIaM~\cite{tian2022generalized}. We demonstrate the great potential of large pretrained ViT-based model on GFSS task based on our experiment results, and show a potential caveat of using pure ViT-based encdoder-decoder architecture in the GFSS setting. 

{
    \small    
    \bibliographystyle{ieeenat_fullname}
    \bibliography{main}
}

\end{document}